\documentclass[11pt,a4paper]{article}
\usepackage[utf8]{inputenc}
\usepackage[T1]{fontenc}
\usepackage{CJKutf8}

\usepackage[margin=1in]{geometry}
\usepackage{microtype}
\usepackage{amsmath,amssymb}

\usepackage{newtxtext,newtxmath}
\usepackage{setspace}
\usepackage[font=small,labelfont=bf,labelsep=period]{caption}
\usepackage{fancyhdr}
\usepackage{titlesec}
\usepackage[shortlabels]{enumitem}
\usepackage[runin]{abstract}
\usepackage[noblocks]{authblk}

\usepackage{graphicx}
\usepackage{booktabs}
\usepackage{multirow}

\usepackage[colorlinks=true, linkcolor=black, citecolor=black, urlcolor=blue!70!black]{hyperref}
\usepackage{natbib}
\let\oldcite\cite
\renewcommand{\cite}[1]{\textit{\oldcite{#1}}}
\let\oldcitep\citep
\renewcommand{\citep}[1]{\textit{\oldcitep{#1}}}

\usepackage{listings}
\usepackage{xcolor}

\usepackage{algorithm}
\usepackage{algpseudocode}

\usepackage{tikz}
\usetikzlibrary{shapes,arrows,positioning,calc,fit}

\definecolor{codegreen}{rgb}{0,0.6,0}
\definecolor{codegray}{rgb}{0.5,0.5,0.5}
\definecolor{codepurple}{rgb}{0.58,0,0.82}
\definecolor{backcolour}{rgb}{0.95,0.95,0.92}

\lstdefinestyle{mystyle}{
    backgroundcolor=\color{backcolour},
    commentstyle=\color{codegreen},
    keywordstyle=\color{magenta},
    numberstyle=\tiny\color{codegray},
    stringstyle=\color{codepurple},
    basicstyle=\ttfamily\footnotesize,
    breakatwhitespace=false,
    breaklines=true,
    captionpos=b,
    keepspaces=true,
    numbers=left,
    numbersep=5pt,
    showspaces=false,
    showstringspaces=false,
    showtabs=false,
    tabsize=2
}
\fancypagestyle{plain}{\fancyhf{}\fancyfoot[C]{\small\thepage}}

\titleformat{\section}{\large\bfseries}{\thesection.}{0.5em}{}
\titleformat{\subsection}{\normalsize\bfseries}{\thesubsection.}{0.5em}{}
\titlespacing*{\section}{0pt}{1.5ex plus 0.5ex}{1ex plus 0.3ex}
\titlespacing*{\subsection}{0pt}{1.2ex plus 0.4ex}{0.8ex plus 0.2ex}

\setlist{nosep,leftmargin=1.5em}
\setlist[itemize]{label=\textbullet}

\title{%
\includegraphics[height=1.5cm]{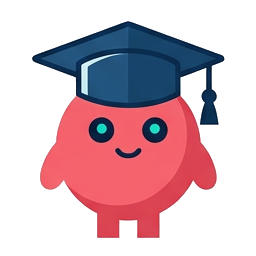}\\[0.3em]%
\Large\bfseries Scaling Laws for Educational AI Agents}

\author[1,2]{Mengsong Wu}
\author[1]{Hao Hao}
\author[2]{Shuzhen Bi}
\author[1]{Keqian Li}
\author[1,2]{Wentao Liu}
\author[1]{Siyu Song}
\author[1]{Hongbo Zhao}
\author[1,2]{Aimin Zhou\textsuperscript{*}}

\affil[ ]{\small\texttt{radi.cat@qq.com, haohao@sjtu.edu.cn, sa22916003@mail.ustc.edu.cn, kqli@mail.ecnu.edu.cn,}}
\affil[ ]{\small\texttt{wtliu@stu.ecnu.edu.cn, siyusong00@gmail.com, hbzhao@stu.ecnu.edu.cn, amzhou@cs.ecnu.cn}}
\affil[1]{East China Normal University}
\affil[2]{Shanghai Innovation Institute}

\date{March 2026\\\smallskip\textit{EduClaw Team}\\[4pt]
{\normalfont\small\textsuperscript{*}Corresponding author}}

\begin{document}
\begin{CJK}{UTF8}{gbsn}

\maketitle
\thispagestyle{plain}


\begin{abstract}
While scaling laws for Large Language Models (LLMs) have been extensively studied along dimensions of model parameters, training data, and compute, the scaling behavior of LLM-based \textit{educational agents} remains unexplored. We propose that educational agent capability scales not merely with the underlying model size, but through structured dimensions that we collectively term the \textbf{Agent Scaling Law}: role definition clarity, skill depth, tool completeness, runtime capability, and educator expertise injection. Central to this framework is \textbf{AgentProfile}, a structured JSON-based specification that serves as the mechanism enabling systematic capability growth of educational agents. We present \textbf{EduClaw}, a profile-driven multi-agent platform that operationalizes this scaling law, demonstrating its effectiveness through the construction and deployment of 330+ educational agent profiles encompassing 1,100+ skill modules across K-12 subjects. Our empirical observations suggest that educational agent performance scales predictably with profile structural richness. We identify two complementary scaling axes---Tool Scaling and Skill Scaling---as future directions, arguing that the path to more capable educational AI lies not solely in larger models, but in stronger structured capability systems.

\end{abstract}

\begin{figure}[htbp]
\centering
\includegraphics[width=\columnwidth]{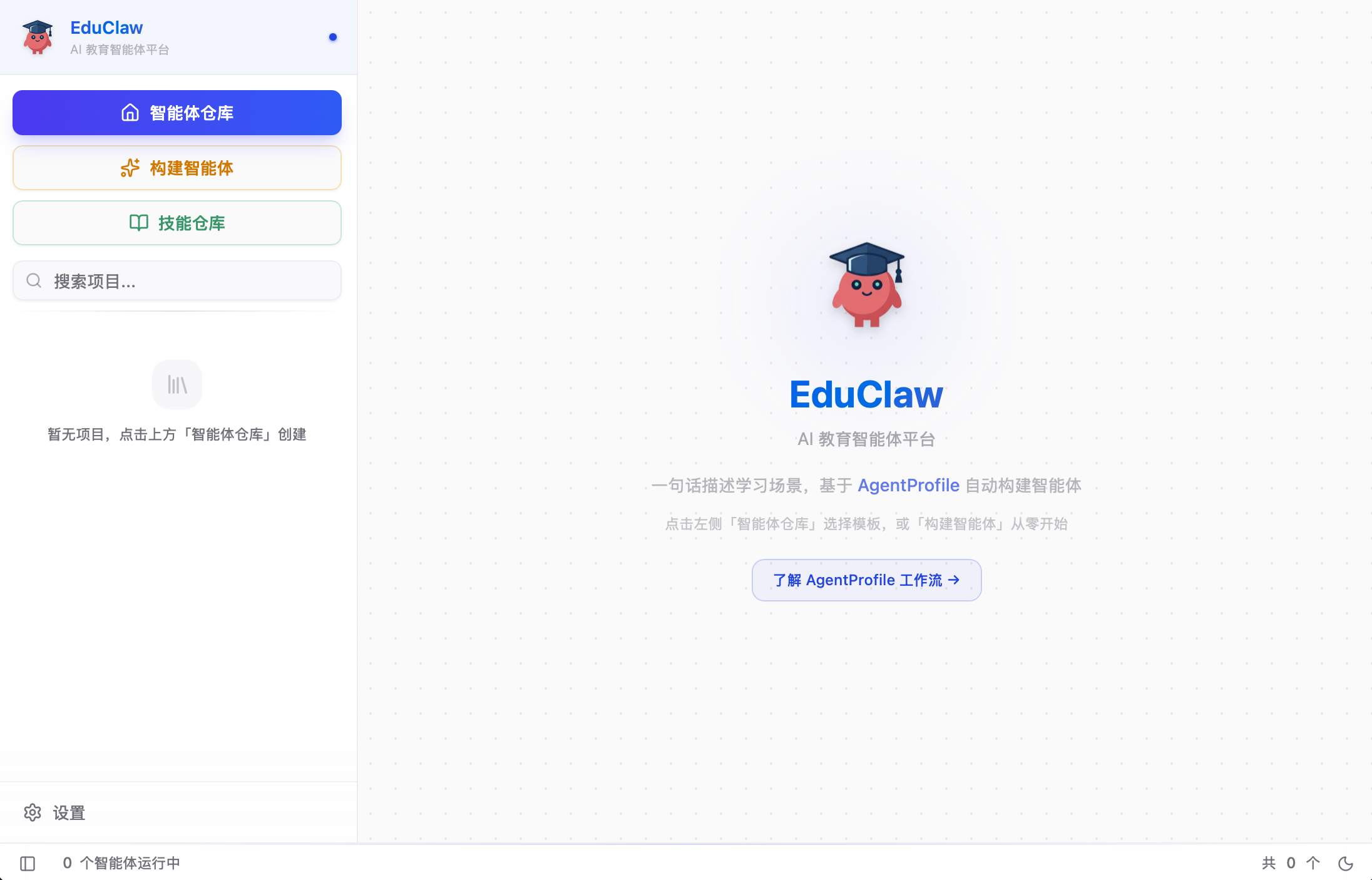}
\caption{EduClaw platform main page. The sidebar provides access to the agent repository, agent construction, and skill repository. The platform supports AgentProfile-based automated agent generation and management.}
\label{fig:mainpage}
\end{figure}


\section{Introduction}\label{sec:introduction}

\subsection{Background}

The discovery of scaling laws has been one of the most consequential findings in modern deep learning. \cite{kaplan2020scaling} demonstrated that language model performance scales as a power law with model parameters, dataset size, and compute budget, while \cite{hoffmann2022training} refined these findings to establish compute-optimal training strategies. These scaling laws have guided the development of foundation models and shaped resource allocation decisions across the field~\citep{brown2020language}.

However, these classical scaling laws describe the behavior of \textit{base models} in isolation. As LLMs are increasingly deployed as components within \textit{agent systems}---particularly in domain-specific applications such as education---a new question arises: \textbf{what governs the scaling of agent-level capabilities?}

\subsection{Motivation}

Educational AI represents a domain where the limitations of model-centric scaling are particularly evident. An LLM with superior benchmark performance does not automatically produce a superior educational agent. Effective tutoring requires pedagogical knowledge, domain-specific strategies, scaffolding behaviors, and alignment with curricular standards---capabilities that emerge not from model size alone, but from the \textit{structured specification} of the agent's role and behavior~\citep{anderson1985intelligent, graesser2004autotutor}.

Consider the difference between asking a general-purpose LLM to ``help with math'' versus deploying an agent with:
\begin{itemize}
    \item A precisely defined role as a middle-school mathematics exploration guide
    \item Structured pedagogical dimensions (divergent thinking, logical rigor, metacognitive monitoring)
    \item Alignment with specific curriculum standards
    \item A repertoire of domain-specific skills and tool integrations
\end{itemize}

The latter consistently produces superior educational interactions---not because of a different underlying model, but because of a richer \textit{capability specification}. This observation motivates our central thesis.

A deeper challenge in educational AI is the \textbf{long-tail distribution of fragmented demands}. Educational needs are extraordinarily diverse: different subjects, grade levels, learning styles, curriculum standards, and pedagogical contexts give rise to a vast combinatorial space of requirements. The traditional approach---training or fine-tuning specialized models for each niche---is neither scalable nor economical. Our key insight is that \textbf{a single foundation model, situated within a well-designed agent environment, can freely explore and generalize across this long-tail distribution}. Rather than encoding every educational scenario into model weights, we externalize the specialization into structured agent profiles, allowing the same underlying model to adapt its behavior through compositional specification. The agent environment serves as an amplifier of the model's generalization capacity, transforming one model into an ecosystem of specialized educational services.

To sustain such an ecosystem at scale, the system must support \textbf{continuous agent evolution}---the ability for agents to be created, refined, composed, and retired as educational needs evolve. This demands a robust underlying architecture that treats agent profiles as first-class, evolvable artifacts rather than static configurations. We argue that agent evolution infrastructure is a prerequisite for realizing educational AI at scale, providing the architectural foundation upon which scaling laws can operate.

\subsection{Our Perspective: Scaling Laws for Educational AI Agents}

We propose that educational agent capability scales along three complementary axes, which we term the \textbf{Educational Agent Scaling Laws}:

\begin{enumerate}
    \item \textbf{Agent Scaling Law}: Agent capability increases with the structural richness of its profile---including role definition clarity, pedagogical dimension depth, skill composition, and multi-agent orchestration. This is operationalized through our \textbf{AgentProfile} specification.
    \item \textbf{Tool Scaling Law}: As the repository of callable tools grows (e.g., equation solvers, diagram generators, assessment rubrics), the agent's actionable capability expands. \textit{(Future work.)}
    \item \textbf{Skill Scaling Law}: As domain-specific skill modules deepen and specialize, the agent's pedagogical expertise scales. \textit{(Future work.)}
\end{enumerate}

This paper focuses on the first axis---the Agent Scaling Law---and provides empirical grounding through the design, implementation, and deployment of the \textbf{EduClaw} platform with OpenClaw\footnote{\url{https://github.com/openclaw/openclaw}} service.

Our ultimate vision is to realize \textbf{Scaling Laws for Educational Services}: just as classical scaling laws have transformed model training by revealing predictable relationships between resources and performance, we aim to establish analogous principles that govern how educational service quality scales with structured investment in agents, tools, and skills. In this paradigm, improving educational outcomes becomes an \textit{engineering problem with predictable returns}---by systematically enriching agent profiles, expanding tool repositories, and deepening skill modules, educational service providers can achieve measurable and compounding improvements in teaching effectiveness. The three scaling axes we propose are the foundation of this vision: when composed together, they define a comprehensive scaling surface for educational AI, enabling the transition from ad hoc prompt engineering to principled, scalable educational service design.

\subsection{Contributions}

This paper makes the following contributions:

\begin{enumerate}
    \item We articulate the \textbf{Agent Scaling Law} for educational AI: the principle that agent capability scales with structured profile richness, not model size alone.
    \item We define the \textbf{AgentProfile} specification, a domain-agnostic open-source protocol standard for structured agent capability definition. While demonstrated in education, AgentProfile serves as a general-purpose agent specification mechanism applicable across diverse domains.
    \item We present the \textbf{EduClaw} platform, a profile-driven multi-agent system that operationalizes the Agent Scaling Law, and demonstrate its effectiveness through the deployment of 330+ educational agents with 1,100+ skill modules across K-12 subjects.
    \item We identify Tool Scaling and Skill Scaling as complementary future research directions and discuss their interaction with agent-level scaling.
\end{enumerate}


\section{Related Work}\label{sec:related}

\subsection{Scaling Laws for Large Language Models}

The study of scaling laws in deep learning has provided foundational insights into model development. \cite{kaplan2020scaling} established that cross-entropy loss of language models follows power-law relationships with model parameters, dataset size, and compute budget, enabling predictable performance improvements through resource scaling. \cite{hoffmann2022training} (Chinchilla) refined these findings, demonstrating that prior models were significantly undertrained relative to their size and proposing compute-optimal training strategies. These works have profoundly influenced the field, but they address scaling at the \textit{model} level---the behavior of a single neural network in isolation.

As LLMs are deployed within agent systems, a gap emerges: classical scaling laws do not account for the structured specifications, tools, and skills that determine agent-level performance. Our work addresses this gap by proposing scaling laws at the \textit{agent} level.

\subsection{LLM-Based Educational AI}

Recent LLM-based educational applications have demonstrated the potential of conversational AI in learning contexts. \cite{khan2023khanmigo} leverages GPT-4 for personalized tutoring across subjects, while \cite{duolingo2023max} applies LLMs to language learning through roleplay and explanation features. These systems represent important advances but typically employ \textit{monolithic} architectures: a single model with a fixed prompt template, offering limited mechanisms for systematic capability growth or domain adaptation.

Traditional intelligent tutoring systems~\citep{anderson1985intelligent, graesser2004autotutor} predate LLMs and rely on hand-crafted cognitive models and dialogue strategies. While these systems encode deep pedagogical knowledge, they lack the flexibility and generative capabilities of modern LLM-based approaches.

\subsection{Agent Frameworks and Profile-Driven Construction}

The concept of structured agent profiles has gained traction in multi-agent systems research. \cite{park2023generative} demonstrated that agents with detailed persona descriptions exhibit more coherent and believable behavior. \cite{wang2023survey} and \cite{xi2023rise} provide comprehensive surveys of LLM-based agent architectures, documenting the shift from prompt-only systems toward structured agent specifications with defined roles, tools, and memory systems.

Multi-agent frameworks for educational contexts have been explored in earlier work~\citep{johnson2000intelligent, biswas2010agent}, though these systems relied on rule-based reasoning and fixed interaction patterns. The integration of modern LLM capabilities with structured agent specifications for education remains largely unexplored.

\subsection{Gap: Scaling Laws for Educational AI Agents}

To our knowledge, no prior work has examined scaling laws at the agent level for educational AI. Existing scaling law research focuses exclusively on model-level properties (parameters, data, compute), while educational AI research focuses on application design without formalizing the relationship between agent specification richness and capability. Our work bridges this gap by proposing the Agent Scaling Law and providing empirical evidence through the AgentProfile framework and EduClaw platform.


\section{Agent Scaling Law via AgentProfile}\label{sec:scaling}

This section presents the core contribution of this paper: the \textbf{Agent Scaling Law} for educational AI, operationalized through the \textbf{AgentProfile} specification. We define the scaling law, describe the AgentProfile mechanism, analyze its scaling dimensions, and present empirical observations.

\subsection{Defining the Agent Scaling Law}

We propose that educational agent capability is a function of multiple structured dimensions, not solely the capacity of the underlying language model:

\begin{equation}\label{eq:scaling}
    C_{\text{agent}} \propto f\!\left(d_{\text{role}},\; d_{\text{dim}},\; d_{\text{skill}},\; d_{\text{tool}},\; d_{\text{runtime}}\right)
\end{equation}

\noindent where:
\begin{itemize}
    \item $C_{\text{agent}}$: Overall educational agent capability
    \item $d_{\text{role}}$: Role definition clarity---precision of pedagogical identity and behavioral specification
    \item $d_{\text{dim}}$: Core dimension depth---richness of structured pedagogical focus areas
    \item $d_{\text{skill}}$: Skill composition---breadth and depth of domain-specific knowledge modules
    \item $d_{\text{tool}}$: Tool completeness---availability of callable actions and integrations
    \item $d_{\text{runtime}}$: Runtime capability---execution environment features (context management, multi-agent coordination)
\end{itemize}

This formulation differs fundamentally from classical scaling laws~\citep{kaplan2020scaling, hoffmann2022training}, which express performance as a function of model parameters $N$, dataset size $D$, and compute $C$. The Agent Scaling Law captures a different level of abstraction: given a fixed base model, how does agent capability grow with the richness of its structured specification?

\subsection{AgentProfile as Scaling Mechanism}

The \textbf{AgentProfile}\footnote{\url{https://github.com/EduClaw-InnoSpark/AgentProfile}} specification is a domain-agnostic, open-source protocol standard for defining AI agent capabilities through structured JSON schemas. While this paper demonstrates its application in education, AgentProfile is designed as a \textit{general-purpose agent specification protocol}: its schema---comprising role definitions, skill modules, tool bindings, and sub-agent orchestration---is equally applicable to domains such as healthcare consultation, legal advisory, customer service, software engineering, and scientific research. The education domain serves as a rigorous proving ground due to its demanding requirements for structured pedagogy, curriculum alignment, and adaptive interaction, but the protocol itself imposes no domain-specific constraints. By establishing AgentProfile as an open standard, we aim to provide the community with a shared foundation for agent interoperability, composability, and systematic capability scaling across arbitrary domains.

In the context of this paper, each profile defines a complete educational agent specification:

\begin{lstlisting}[caption={AgentProfile Schema Template (Version 1.0)}, label={lst:profile}, basicstyle=\ttfamily\footnotesize, breaklines=true, numbers=left, backgroundcolor=\color{backcolour}]
{
  "name":           "<agent_name>",
  "description":    "<one_sentence_purpose>",
  "details":        "<structured_markdown>",
  "agent_template": "<template_id>",
  "skills":         ["<skill_id_1>", "<skill_id_2>", ...],
  "tools":          ["<tool_id_1>", "<tool_id_2>", ...],
  "subagents":      ["<subagent_id_1>", ...]
}
\end{lstlisting}

Table~\ref{tab:profile-fields} describes each field and its role in agent capability scaling.

\begin{table}[t]
\centering
\small
\caption{AgentProfile Schema Fields}
\label{tab:profile-fields}
\begin{tabular}{@{}llp{5.5cm}@{}}
\toprule
\textbf{Field} & \textbf{Type} & \textbf{Description} \\
\midrule
\texttt{name} & String & Concise identifier reflecting the agent's role \\
\texttt{description} & String & One-sentence summary of the agent's purpose \\
\texttt{details} & Markdown & Structured behavioral spec with four sections (Section~\ref{sec:details-format}) \\
\texttt{agent\_template} & String & Base runtime template and default configs \\
\texttt{skills} & List & Skill module references from the repository \\
\texttt{tools} & List & Tool integrations (solvers, generators, etc.) \\
\texttt{subagents} & List & Subordinate agents for task decomposition \\
\bottomrule
\end{tabular}
\end{table}

\subsubsection{Details Format Specification}\label{sec:details-format}

The \texttt{details} field employs a structured Markdown format with four mandatory sections, each contributing a distinct scaling dimension:

\paragraph{Role Definition.} Defines the agent's pedagogical identity, interaction tone, and philosophical approach:

\begin{quote}
\textit{As a [domain] [function] assistant, use a [style] approach to help users [core goal]. Focus on [key principle].}
\end{quote}

\paragraph{Core Dimensions.} A structured breakdown of pedagogical focus areas using tabular format. Table~\ref{tab:dimensions} illustrates an example for mathematics tutoring.

\begin{table}[t]
\centering
\small
\caption{Example Core Dimensions for Mathematics Tutoring}
\label{tab:dimensions}
\begin{tabular}{@{}lp{6cm}@{}}
\toprule
\textbf{Dimension} & \textbf{Focus Points} \\
\midrule
Divergent Thinking & Path diversity, cross-domain associations \\
Logical Rigor & Reasoning completeness, counterexample construction \\
Math Expression & Symbolic normativity, geometry-algebra translation \\
Inquiry Depth & Problem variation, essential pattern extraction \\
Metacognition & Strategy evaluation, obstacle diagnosis \\
\bottomrule
\end{tabular}
\end{table}

\paragraph{Standards.} Quality criteria and reference frameworks ensuring curriculum alignment:

\begin{itemize}
    \item Curriculum standards (e.g., national mathematics curriculum standards)
    \item Assessment rubrics and evaluation methodologies
    \item Pedagogical principles (scaffolding, zone of proximal development~\citep{vygotsky1978mind})
\end{itemize}

\paragraph{Output Format.} Structured response templates ensuring consistent pedagogical delivery:

\begin{enumerate}
    \item \textbf{Problem Deconstruction}: Context analysis and condition mapping
    \item \textbf{Thinking Activation}: Multi-directional heuristic prompts
    \item \textbf{Path Exploration}: Process accompaniment with obstacle diagnosis
    \item \textbf{Solution Comparison}: Structured comparison of multiple approaches
    \item \textbf{Variation Extension}: Progressive problem chain design
    \item \textbf{Inquiry Log}: Metacognitive reflection prompts
\end{enumerate}

\subsection{Scaling Dimensions}\label{sec:dimensions}

We identify four primary dimensions along which the Agent Scaling Law operates:

\subsubsection{Role Definition Clarity}

A more precisely defined role produces more consistent and pedagogically appropriate behavior. A minimal role definition (``math tutor'') yields generic responses, while a richly specified role (``middle-school mathematics exploration guide using Socratic questioning to develop divergent thinking and mathematical modeling ability'') produces targeted, pedagogically grounded interactions. Role clarity scales the agent's behavioral coherence and domain appropriateness.

\subsubsection{Core Dimension Depth}

The core dimensions table provides a structured decomposition of pedagogical focus areas. Adding dimensions (e.g., metacognitive monitoring, assessment alignment) broadens the agent's pedagogical coverage, while deepening focus points within each dimension enhances specificity. This creates a two-dimensional scaling surface: breadth (number of dimensions) and depth (detail per dimension).

\subsubsection{Skill Composition}

Skills are modular, reusable knowledge units that encode domain expertise. Each skill module includes behavioral specifications, applicable scenarios, guided principles, and output templates. Agent capability scales with:
\begin{itemize}
    \item \textbf{Skill count}: More skills enable coverage of more educational scenarios
    \item \textbf{Skill quality}: Better-structured skills produce more effective pedagogical interactions
    \item \textbf{Skill composition}: Combinations of complementary skills enable emergent capabilities
\end{itemize}

\subsubsection{Multi-Agent Orchestration}

The \texttt{subagents} field enables hierarchical task decomposition. A primary agent can delegate specialized tasks (e.g., equation solving, diagram explanation, assessment generation) to subordinate agents, each with their own profile. This enables capability scaling through \textit{composition}: the orchestrating agent's effective capability is greater than the sum of its components due to coordinated task distribution.

\subsection{End-to-End Pipeline: From One Sentence to Running Agent}\label{sec:constructor}

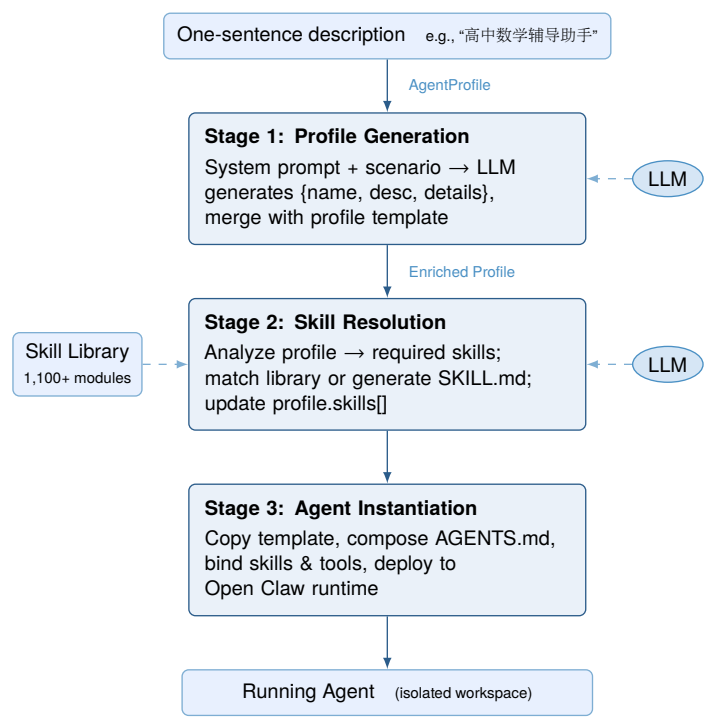
\begin{figure}[t]
\centering
\definecolor{clrDark}{HTML}{2B5C8A}
\definecolor{clrMid}{HTML}{4A90C4}
\definecolor{clrLight}{HTML}{D6E6F5}
\definecolor{clrPale}{HTML}{EAF1F9}
\definecolor{clrDash}{HTML}{7EAED3}
\definecolor{clrAccent}{HTML}{E8F0FE}
\begin{tikzpicture}[
    >=latex,
    stagebox/.style={
        rectangle, draw=clrDark, line width=0.5pt,
        rounded corners=2.5pt, align=left, inner sep=6pt,
        fill=clrPale, font=\sffamily\scriptsize,
        text width=4.8cm
    },
    io/.style={
        rectangle, draw=clrDash, line width=0.5pt,
        rounded corners=2.5pt, align=center, inner sep=5pt,
        fill=clrAccent, font=\sffamily\scriptsize,
        minimum width=5.4cm
    },
    llm/.style={
        ellipse, draw=clrMid, line width=0.5pt,
        align=center, inner sep=2pt,
        fill=clrLight, font=\sffamily\scriptsize
    },
    sideio/.style={
        rectangle, draw=clrDash, line width=0.5pt,
        rounded corners=2.5pt, align=center, inner sep=4pt,
        fill=clrAccent, font=\sffamily\scriptsize
    },
    arr/.style={->, draw=clrDark, line width=0.5pt},
    darr/.style={->, draw=clrDash, line width=0.5pt, dashed},
    stlabel/.style={font=\sffamily\scriptsize\bfseries, text=clrDark},
]

\node[io] (input) {One-sentence description\quad{\tiny e.g., ``高中数学辅导助手''}};

\node[stagebox, below=0.7cm of input] (s1) {
    \textbf{Stage 1: Profile Generation}\\[2pt]
    System prompt + scenario $\rightarrow$ LLM\\
    generates \{name, desc, details\},\\
    merge with profile template
};
\node[llm, right=0.6cm of s1] (llm1) {LLM};
\draw[darr] (llm1) -- (s1);

\node[font=\sffamily\tiny, text=clrMid, right=0.15cm]
    at ($(input.south)!0.5!(s1.north)$) {AgentProfile};

\node[stagebox, below=0.7cm of s1] (s2) {
    \textbf{Stage 2: Skill Resolution}\\[2pt]
    Analyze profile $\rightarrow$ required skills;\\
    match library or generate SKILL.md;\\
    update profile.skills[]
};
\node[llm, right=0.6cm of s2] (llm2) {LLM};
\draw[darr] (llm2) -- (s2);
\node[sideio, left=0.6cm of s2] (lib) {Skill Library\\{\tiny 1,100+ modules}};
\draw[darr] (lib) -- (s2);

\node[font=\sffamily\tiny, text=clrMid, right=0.15cm]
    at ($(s1.south)!0.5!(s2.north)$) {Enriched Profile};

\node[stagebox, below=0.7cm of s2] (s3) {
    \textbf{Stage 3: Agent Instantiation}\\[2pt]
    Copy template, compose AGENTS.md,\\
    bind skills \& tools, deploy to\\
    Open Claw runtime
};

\node[io, below=0.7cm of s3] (output) {Running Agent\quad{\tiny (isolated workspace)}};

\draw[arr] (input) -- (s1);
\draw[arr] (s1) -- (s2);
\draw[arr] (s2) -- (s3);
\draw[arr] (s3) -- (output);

\end{tikzpicture}
\caption{End-to-end pipeline from one-sentence description to running educational agent. Stages 1 and 2 are LLM-powered (dashed arrows); Stage 3 is deterministic construction.}
\label{fig:pipeline}
\end{figure}

The Agent Scaling Law is operationalized through a three-stage pipeline (Figure~\ref{fig:pipeline}) that transforms a single natural language sentence into a fully functional educational agent. Each stage progressively enriches the agent's structured specification, directly contributing to its capability scaling.

\subsubsection{Stage 1: Profile Generation (LLM-Powered)}

Given a one-sentence teaching scenario description (e.g., ``high school mathematics tutoring assistant''), an LLM generates a complete \textbf{AgentProfile} in a single call. The generation is guided by a system prompt that specifies the required JSON structure and provides a reference example of the four-section \texttt{details} format (Role Definition, Core Dimensions, Standards, Output Format). The LLM output is parsed, validated, and merged with a base profile template that provides default values for \texttt{agent\_template}, \texttt{tools}, and \texttt{subagents}.

This stage is the entry point for capability scaling: the quality of the generated profile---its role clarity, dimension richness, and standard specificity---directly determines the agent's baseline capability, as formalized in Equation~\ref{eq:scaling}.

\subsubsection{Stage 2: Skill Resolution (LLM-Powered Matching)}

The generated profile initially has an empty \texttt{skills} array. In this stage, a two-step process enriches the profile with domain-specific skill modules:

\begin{enumerate}
    \item \textbf{Skill Analysis}: The LLM analyzes the profile (name, description, details) and identifies a set of required skill modules, returning a list of skill identifiers.
    \item \textbf{Matching \& Generation}: Each required skill is matched against the existing skill library (1,100+ modules). For matched skills, references are added directly to the profile. For missing skills, the LLM generates a complete \texttt{SKILL.md} specification---including applicable scenarios, pedagogical dimensions, guiding principles, and output format templates---which is then added to the library for future reuse.
\end{enumerate}

This stage implements the \textit{Skill Composition} scaling dimension (Section~\ref{sec:dimensions}): skill count, quality, and complementarity collectively determine the agent's domain expertise depth. The library grows monotonically, creating a positive feedback loop where each new agent can benefit from skills generated for previous agents.

\subsubsection{Stage 3: Agent Instantiation (Deterministic Construction)}

The enriched AgentProfile is transformed into a runnable agent through Algorithm~\ref{alg:construction}:

\begin{algorithm}[t]
\caption{Agent Construction from AgentProfile}
\label{alg:construction}
\begin{algorithmic}[1]
\Require AgentProfile $P = (\textit{name}, \textit{desc}, \textit{details}, \textit{skills}, \textit{tools}, \textit{subagents})$
\Require Skill library $\mathcal{S}$, Tool registry $\mathcal{T}$, Agent registry $\mathcal{A}$
\Ensure Runnable agent instance $\alpha$
\Statex \textit{// Phase 1: Profile Resolution}
\State $(\textit{role},\, \textit{dims},\, \textit{stds},\, \textit{fmt}) \gets \textsc{ParseDetails}(P.\textit{details})$
\Statex \textit{// Phase 2: Capability Assembly}
\State $\mathcal{S}_P \gets \{\textsc{Resolve}(s,\, \mathcal{S}) \mid s \in P.\textit{skills}\}$ \Comment{Bind skills}
\State $\mathcal{T}_P \gets \{\textsc{Resolve}(t,\, \mathcal{T}) \mid t \in P.\textit{tools}\}$ \Comment{Bind tools}
\State $\mathcal{A}_P \gets \{\textsc{Construct}(\textsc{Resolve}(a,\, \mathcal{A})) \mid a \in P.\textit{subagents}\}$ \Comment{Recursive}
\Statex \textit{// Phase 3: Agent Instantiation}
\State $\textit{spec} \gets \textsc{Compose}(\textit{role},\, \textit{dims},\, \textit{stds},\, \textit{fmt},\, \mathcal{S}_P)$
\State $\alpha \gets \textsc{Instantiate}(\textit{spec},\, \mathcal{T}_P,\, \mathcal{A}_P)$
\State \Return $\alpha$
\end{algorithmic}
\end{algorithm}

Concretely, this phase: (1)~copies a base agent template to create an isolated workspace; (2)~composes a behavioral specification (\texttt{AGENTS.md}) from the profile's structured fields; (3)~copies referenced skill modules from the library into the workspace; (4)~registers tool interfaces and recursively constructs any declared subagents; and (5)~deploys the agent as an isolated Open Claw runtime instance with its own process, context, and communication endpoints.

\begin{figure}[b]
\centering
\includegraphics[width=\columnwidth]{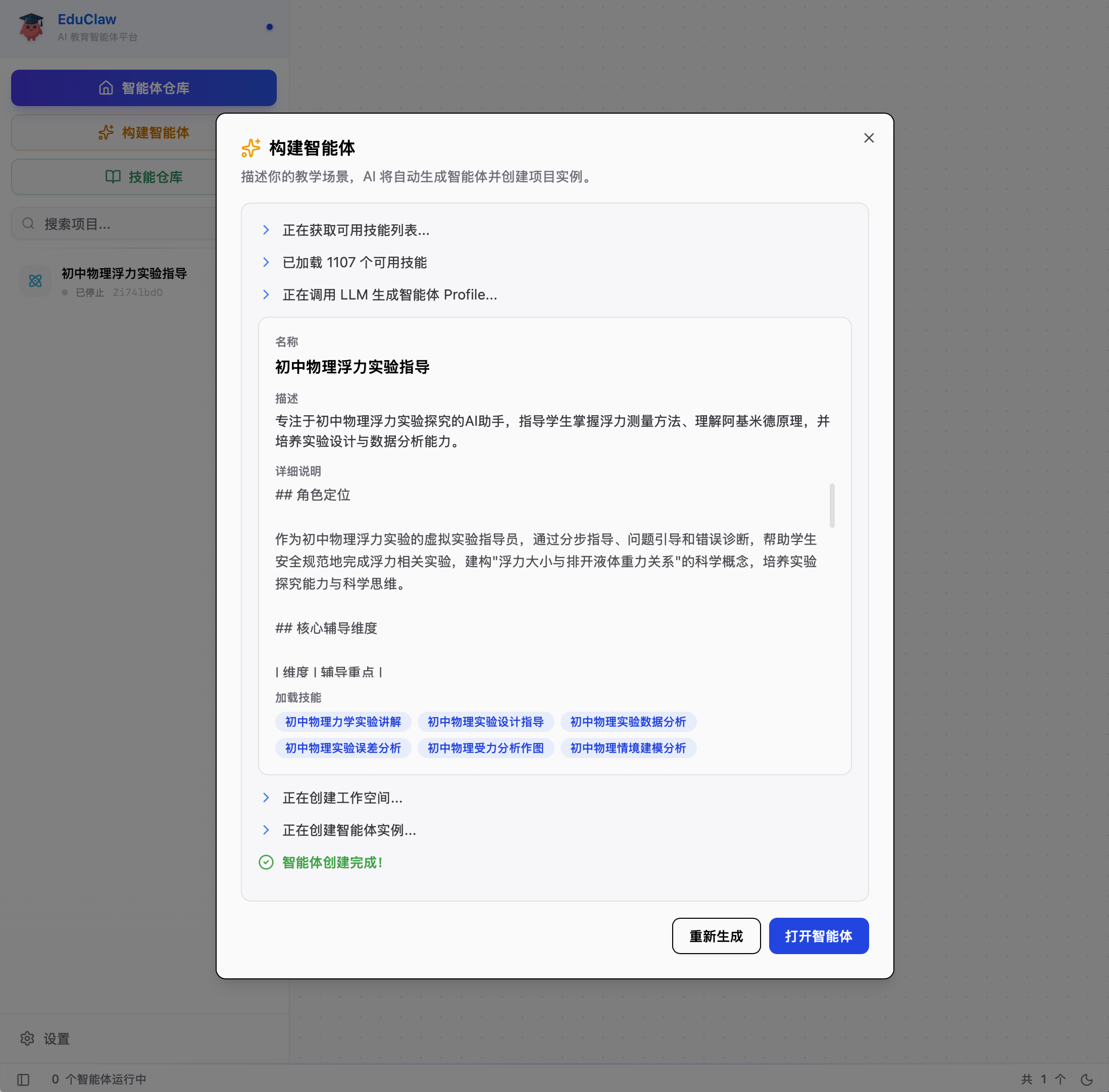}
\caption{EduClaw agent construction interface. From a one-sentence input, the system generates an AgentProfile, matches skills from the library (shown as tags), and produces a running agent.}
\label{fig:construct}
\end{figure}

Subagent construction is \textit{recursive}: each subagent referenced in the profile is itself constructed via the same algorithm, enabling hierarchical agent topologies of arbitrary depth. The entire pipeline from one-sentence input to running agent completes in under one minute, enabling rapid iteration on agent specifications---a key practical enabler of the scaling law. Figure~\ref{fig:construct} shows this pipeline in action within the EduClaw interface.

\subsection{Empirical Observations}\label{sec:observations}

Through the development and deployment of 330+ agent profiles across K-12 subjects, we observe several patterns consistent with the Agent Scaling Law:

\begin{enumerate}
    \item \textbf{Profile richness correlates with interaction quality}: Agents with more detailed role definitions, more core dimensions, and richer output format specifications consistently produce more pedagogically appropriate and contextually relevant responses.
    \item \textbf{Skill composition enables specialization}: Agents equipped with domain-specific skill modules demonstrate markedly superior performance in their target domains compared to profile-only agents.
    \item \textbf{Diminishing returns at extremes}: Excessively detailed profiles can overwhelm the context window, suggesting an optimal profile complexity that balances specification richness with model capacity---analogous to the compute-optimal balance identified by~\citep{hoffmann2022training}.
    \item \textbf{Cross-subject transfer}: Well-structured profiles in one subject area serve as effective templates for related subjects, suggesting that the scaling mechanism generalizes across domains.
\end{enumerate}


\section{EduClaw Platform}\label{sec:system}

The Agent Scaling Law described in Section~\ref{sec:scaling} requires a platform capable of constructing, managing, and executing profile-driven agents at scale. This section presents the \textbf{EduClaw} platform architecture that operationalizes the scaling law.

\subsection{Architecture Overview}

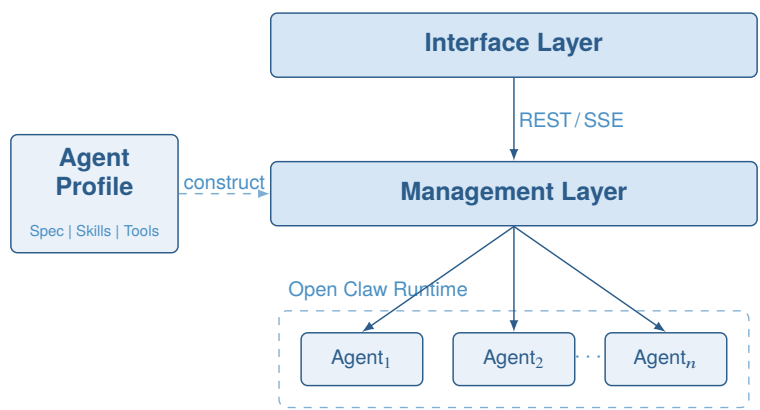
\begin{figure}[htbp]
\centering
\definecolor{clrDark}{HTML}{2B5C8A}    
\definecolor{clrMid}{HTML}{4A90C4}     
\definecolor{clrLight}{HTML}{D6E6F5}   
\definecolor{clrPale}{HTML}{EAF1F9}    
\definecolor{clrDash}{HTML}{7EAED3}    

\begin{tikzpicture}[
    >=latex,
    box/.style={
        rectangle, draw=clrDark, line width=0.55pt,
        rounded corners=2.5pt, align=center,
        inner sep=5pt
    },
]

\def\layerw{6.4cm}
\def\layerh{0.85cm}
\def\agentw{1.6cm}
\def\agenth{0.7cm}

\node[box, minimum width=\layerw, minimum height=\layerh, fill=clrLight]
    (L1) {{\sffamily\small\bfseries\color{clrDark} Interface Layer}};

\node[box, minimum width=\layerw, minimum height=\layerh, fill=clrLight,
    below=1.1cm of L1]
    (L2) {{\sffamily\small\bfseries\color{clrDark} Management Layer}};

\node[box, minimum width=\agentw, minimum height=\agenth, fill=clrPale,
    below=1.4cm of L2, xshift=-2cm]
    (A1) {{\sffamily\scriptsize\color{clrDark} Agent$_1$}};
\node[box, minimum width=\agentw, minimum height=\agenth, fill=clrPale,
    below=1.4cm of L2]
    (A2) {{\sffamily\scriptsize\color{clrDark} Agent$_2$}};
\node[box, minimum width=\agentw, minimum height=\agenth, fill=clrPale,
    below=1.4cm of L2, xshift=2cm]
    (An) {{\sffamily\scriptsize\color{clrDark} Agent$_n$}};

\path (A2) -- (An) node[midway, text=clrMid] {{\scriptsize$\cdots$}};

\node[rectangle, draw=clrDash, line width=0.5pt, rounded corners=3pt,
    dashed, inner xsep=8pt, inner ysep=8pt, fit=(A1)(A2)(An)] (Abox) {};
\node[anchor=south west, font=\sffamily\scriptsize, text=clrMid]
    at (Abox.north west) {Open Claw Runtime};

\node[box, minimum width=2.2cm, minimum height=1.4cm, fill=clrPale,
    left=1.2cm of L2]
    (P) {{\sffamily\small\bfseries\color{clrDark}\shortstack{Agent\\[-1pt]Profile}}\\[2pt]
         {\sffamily\tiny\color{clrMid} Spec\;\textbar\;Skills\;\textbar\;Tools}};

\draw[->, draw=clrDark, line width=0.55pt]
    (L1.south) -- node[right, font=\sffamily\scriptsize, text=clrMid, inner sep=1.5pt]
    {REST\,/\,SSE} (L2.north);

\draw[->, draw=clrDark, line width=0.55pt] (L2.south) -- (A1.north);
\draw[->, draw=clrDark, line width=0.55pt] (L2.south) -- (A2.north);
\draw[->, draw=clrDark, line width=0.55pt] (L2.south) -- (An.north);

\draw[->, draw=clrDash, line width=0.55pt, dashed]
    (P.east) -- node[above, font=\sffamily\scriptsize, text=clrMid, inner sep=1.5pt]
    {construct} (L2.west);

\end{tikzpicture}
\caption{EduClaw three-tier architecture. AgentProfiles provide declarative specifications that drive agent construction; the management layer orchestrates process lifecycle and request routing; each agent runs as an isolated Open Claw runtime instance.}
\label{fig:architecture}
\end{figure}

EduClaw employs a three-tier architecture (Figure~\ref{fig:architecture}) designed to support scalable deployment of educational agents while maintaining process isolation and pedagogical consistency.

The architecture consists of three layers:

\begin{itemize}
    \item \textbf{Interface Layer}: A web-based interface providing multi-tab conversational interactions with agents, supporting concurrent sessions and administrative controls.
    \item \textbf{Management Layer}: A Node.js server handling agent lifecycle management, API proxying, and Server-Sent Events (SSE) aggregation across all active agent instances.
    \item \textbf{Agent Layer (Open Claw Runtime)}: Isolated Open Claw agent processes, each representing an educational agent with a dedicated workspace and configuration derived from its AgentProfile. Open Claw handles context management, tool orchestration, multi-turn dialogue control, and sub-agent collaboration.
\end{itemize}

\subsection{Agent Construction Pipeline}

The construction pipeline implements the three-phase process formalized in Algorithm~\ref{alg:construction}. Given an AgentProfile, the pipeline proceeds as follows:

\begin{enumerate}
    \item \textbf{Profile Resolution}: The profile's \texttt{details} field is parsed into its structured components---role definition, core dimensions, standards, and output format---which together form the agent's behavioral specification.
    \item \textbf{Capability Assembly}: Declared skills are resolved from the skill library, tool interfaces are bound from the tool registry, and any referenced subagents are recursively constructed, yielding the agent's full capability surface.
    \item \textbf{Agent Instantiation}: All resolved components are composed into a unified specification and deployed as an isolated Open Claw runtime instance with its own workspace, context, and communication endpoints.
\end{enumerate}

Each constructed agent instance encapsulates:
\begin{itemize}
    \item A behavioral specification synthesized from the profile's structured fields
    \item Bound skill modules providing domain-specific pedagogical knowledge
    \item Registered tool interfaces enabling external action execution
    \item References to subordinate agents for hierarchical task delegation
\end{itemize}

\subsection{Process Lifecycle Management}

The management layer handles the complete agent process lifecycle:

\begin{itemize}
    \item \textbf{Spawn}: Launch an Open Claw agent process with a dynamically allocated port for each constructed agent
    \item \textbf{Health Check}: Monitor process status via HTTP health endpoints, with automatic restart on failure
    \item \textbf{Proxy}: Route API requests from the interface layer to the appropriate agent instance based on session context
    \item \textbf{SSE Aggregation}: Collect and multiplex Server-Sent Events from all running agents into a unified event stream
    \item \textbf{Cleanup}: Graceful shutdown on session termination, with resource reclamation
\end{itemize}

\subsection{Open Claw Runtime and Scaling}

Each educational agent runs on \textbf{Open Claw}, an open agent runtime responsible for loading AgentProfiles and skill modules, managing execution context, orchestrating tool calls, and coordinating sub-agent collaboration. The platform supports runtime scaling through several mechanisms:

\begin{itemize}
    \item \textbf{Dynamic Port Allocation}: Each agent instance receives a unique port, enabling concurrent execution without conflicts
    \item \textbf{Idle Auto-Shutdown}: Inactive agent processes are automatically terminated after a configurable timeout, freeing system resources
    \item \textbf{Process Isolation}: Each agent runs in its own process with a dedicated workspace, preventing cross-contamination of context between educational sessions
    \item \textbf{On-Demand Instantiation}: Agents are constructed and launched only when needed, enabling the platform to support a large profile library without proportional resource consumption
\end{itemize}


\section{Empirical Evidence}\label{sec:evaluation}

This section presents empirical evidence supporting the Agent Scaling Law through the scale of deployment, subject coverage analysis, profile quality observations, and educational design principles embedded in the agent profiles.

\subsection{Scale of Deployment}

The EduClaw platform has been used to construct and deploy educational agents at significant scale:

\begin{itemize}
    \item \textbf{330+ Agent Profiles}: Covering K-12 subjects and grade levels, each defined through the AgentProfile specification
    \item \textbf{1,100+ Skill Modules}: Reusable pedagogical components in the skill repository, referenced by agent profiles
    \item \textbf{Sub-Minute Instantiation}: Agent creation from profile to running process in under one minute
\end{itemize}

This scale enables meaningful observation of patterns in the relationship between profile structure and agent capability. Critically, the construction of educational tasks and the skill module library was carried out with the guidance, validation, and endorsement of authoritative education experts, including experienced K-12 teachers, curriculum designers, and educational researchers. These domain experts contributed to defining pedagogical standards, reviewing skill module quality, validating curriculum alignment, and ensuring that agent behaviors conform to established educational best practices. Their involvement ensures that the skill repository reflects expert-level pedagogical knowledge rather than purely model-generated content, lending professional credibility and practical grounding to the platform's educational capabilities.

\begin{figure}[t]
\centering
\includegraphics[width=\columnwidth]{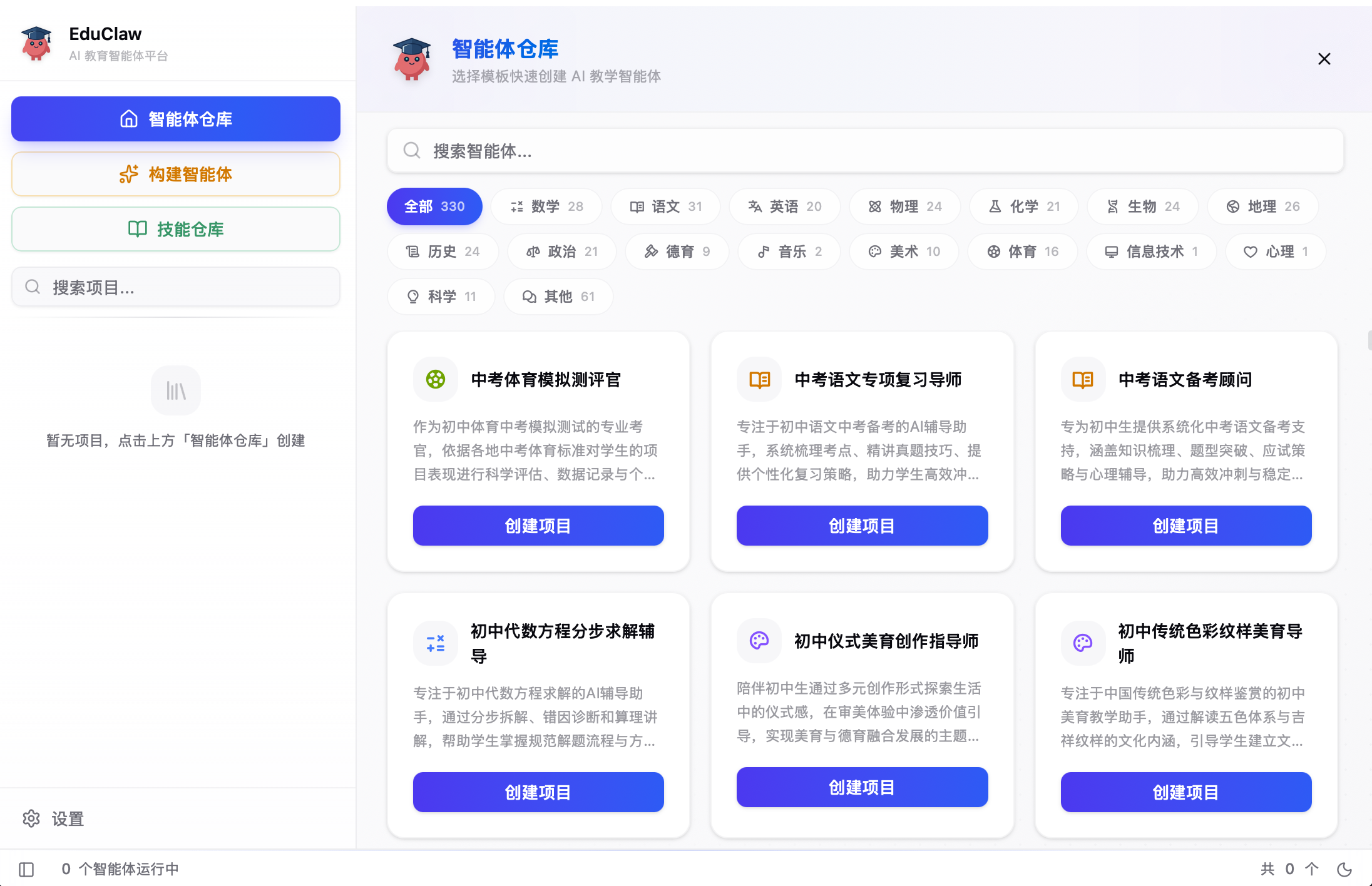}
\caption{EduClaw agent repository. Agents are organized by subject and filterable by grade level.}
\label{fig:agent-repo}
\end{figure}

Figure~\ref{fig:agent-repo} shows the agent repository interface, where agents are organized by subject and grade level, enabling educators to browse and instantiate agents on demand.

\subsection{Subject Coverage}

Table~\ref{tab:coverage} summarizes the distribution of skill modules across subjects and grade levels:

\begin{table}[b]
\centering
\small
\caption{Educational Skill Module Distribution by Subject and Level}
\label{tab:coverage}
\begin{tabular}{@{}lccc@{}}
\toprule
\textbf{Subject} & \textbf{Primary} & \textbf{Middle} & \textbf{High} \\
\midrule
Mathematics & 45 & 68 & 52 \\
Chinese Language & 38 & 42 & 35 \\
English & 32 & 45 & 48 \\
Physics & 12 & 28 & 41 \\
Chemistry & -- & 18 & 35 \\
Biology & 15 & 22 & 38 \\
History & 18 & 25 & 30 \\
Geography & 14 & 20 & 28 \\
Physical Education & 22 & 18 & 12 \\
\midrule
\textbf{Total} & 196 & 286 & 319 \\
\bottomrule
\end{tabular}
\end{table}

The distribution reveals that STEM subjects at the middle and high school levels have the densest coverage, reflecting both curriculum complexity and the availability of structured pedagogical strategies. The breadth of coverage across nine subjects and three grade bands demonstrates the generality of the AgentProfile specification as a scaling mechanism.

\subsection{Profile Quality Analysis}

Analysis of the 330+ deployed profiles reveals structural patterns consistent with the Agent Scaling Law:

\begin{enumerate}
    \item \textbf{Role definition specificity}: Profiles with more specific role definitions (averaging 50+ words in the role section) produce agents with more consistent pedagogical behavior compared to profiles with generic descriptions (under 20 words).
    \item \textbf{Dimension count}: Profiles typically contain 3--7 core dimensions. Profiles with 5+ dimensions show broader pedagogical coverage, though profiles exceeding 7 dimensions show diminishing returns in practice.
    \item \textbf{Skill attachment}: Profiles referencing 2--4 complementary skills demonstrate stronger domain specialization than single-skill profiles, suggesting compositional scaling effects.
    \item \textbf{Output format structure}: Profiles with detailed output format specifications (4+ structured stages) produce more pedagogically consistent responses than those with minimal or absent output templates.
\end{enumerate}

\begin{figure}[t]
\centering
\includegraphics[width=\columnwidth]{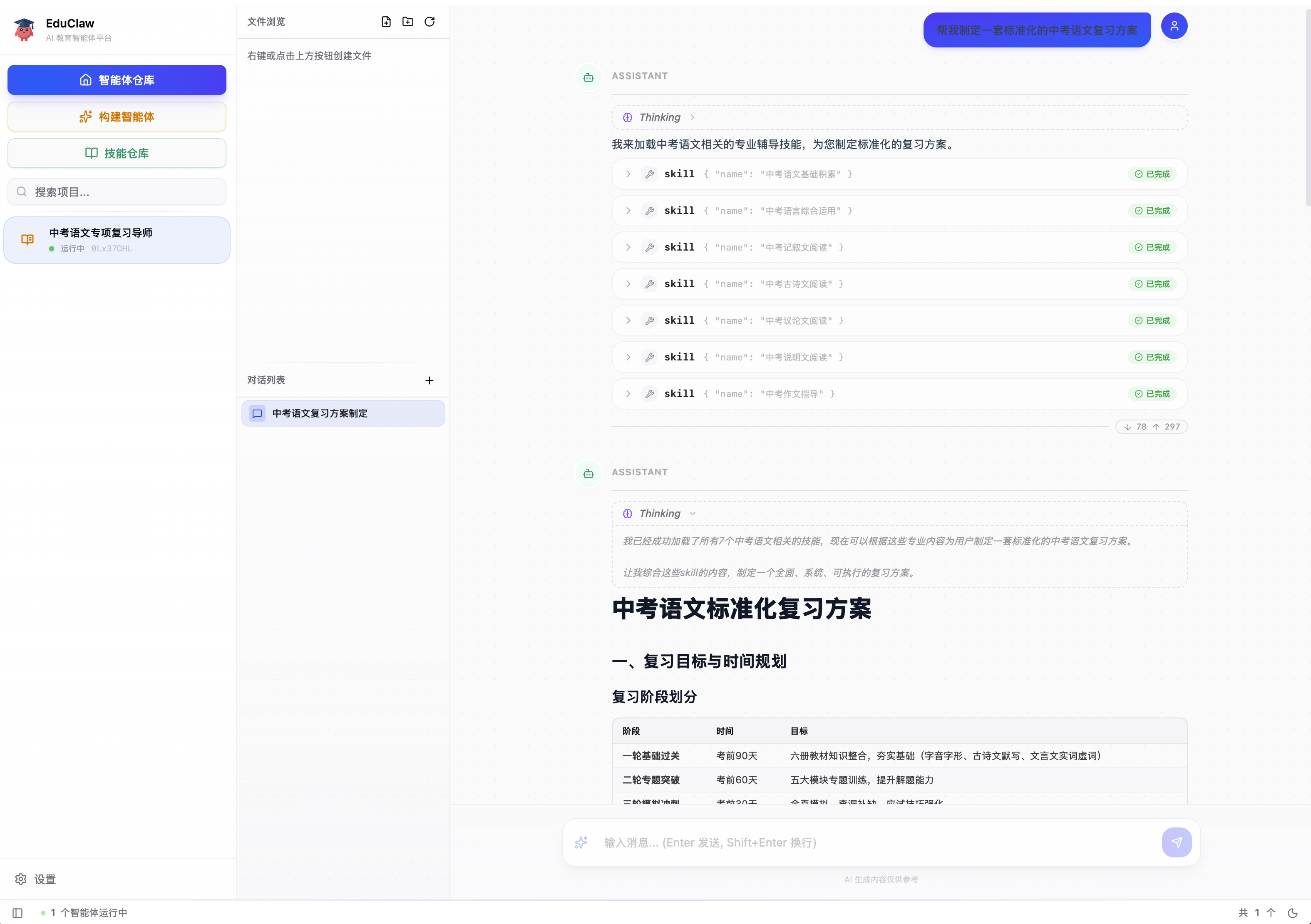}
\caption{An educational agent in action. The agent exhibits structured reasoning (thinking block), recommends relevant skills, and engages in multi-turn pedagogical dialogue.}
\label{fig:testcase}
\end{figure}

Figure~\ref{fig:testcase} illustrates a deployed agent in action: a Chinese language review agent demonstrates structured reasoning, domain-specific skill invocation, and pedagogically appropriate multi-turn interaction---behaviors that emerge from the richness of its AgentProfile specification.

\subsection{Educational Design Principles}

The AgentProfile specification embeds educational design principles that contribute to agent effectiveness. Rather than relying solely on the base model's implicit pedagogical knowledge, profiles \textit{explicitly encode} research-backed instructional strategies as structured behavioral specifications. This section details how key principles from educational theory are operationalized within the profile framework.

\subsubsection{Scaffolding and Zone of Proximal Development}

Following \cite{vygotsky1978mind}'s Zone of Proximal Development (ZPD) theory and \cite{wood1976role}'s scaffolding framework, agent profiles encode multi-level support strategies that adapt to the learner's current capability. The core insight is that effective tutoring operates in the gap between what a student can do independently and what they can achieve with guidance. Profiles operationalize this through three mechanisms:

\begin{itemize}
    \item \textbf{Progressive Support (Hint--Assist--Release)}: Profiles specify a graduated intervention sequence. The agent first offers indirect hints (e.g., ``What theorem relates the sides of a right triangle?''), then provides structured assistance if the student remains stuck (e.g., presenting the Pythagorean theorem with a labeled diagram), and finally releases the student to solve independently. The output format section of the profile encodes the transition conditions between levels.
    \item \textbf{Metacognitive Prompts}: Drawing on \cite{flavell1979metacognition}'s metacognition framework, profiles include explicit instructions for the agent to ask reflective questions such as ``What strategy did you use?'', ``Why did you choose this approach?'', and ``How would you verify your answer?''. These prompts are embedded in the output format as a mandatory \textit{Inquiry Log} stage (see Section~\ref{sec:details-format}), ensuring that every interaction cycle includes a metacognitive component.
    \item \textbf{Error as Learning Resource}: Rather than simply correcting mistakes, profiles instruct agents to treat errors diagnostically---identifying the underlying misconception, presenting a targeted counterexample, and guiding the student to self-correct. For example, a mathematics profile specifies: ``When the student makes an error, do not provide the correct answer immediately. Instead, ask the student to check their work by substituting the result back into the original equation.''
    \item \textbf{Adaptive Difficulty Calibration}: Profiles encode rules for dynamically adjusting problem difficulty based on student performance patterns. When a student answers correctly with confidence, the agent escalates to higher-order questions on Bloom's taxonomy~\citep{bloom1956taxonomy}; when the student struggles, the agent decomposes the problem into smaller sub-tasks, effectively narrowing the ZPD window.
\end{itemize}

This scaffolding approach resonates with \cite{bloom1984two}'s finding that one-on-one tutoring produces a two-sigma improvement over conventional instruction---the structured profile effectively transforms a general-purpose LLM into a personalized tutor that approximates this benefit at scale.

\subsubsection{Multiple Solution Pathways and Divergent Thinking}

For STEM subjects, profiles structure agents to encourage divergent thinking and mathematical creativity, following \cite{polya1945solve}'s problem-solving heuristics. The \textit{Core Dimensions} section of profiles (Table~\ref{tab:dimensions}) typically includes a ``Divergent Thinking'' dimension that operationalizes this principle through a structured pedagogical sequence:

\begin{enumerate}
    \item \textbf{Approach Elicitation}: Before presenting solutions, the agent asks the student to brainstorm possible approaches, developing autonomous problem-solving habits. The profile specifies: ``Always ask the student for their initial approach before offering guidance.''
    \item \textbf{Multi-Path Presentation}: The agent presents 3--5 different solution directions (e.g., algebraic manipulation, geometric interpretation, coordinate methods, special case analysis, proof by contradiction) and encourages the student to explore at least two.
    \item \textbf{Just-in-Time Assistance}: As the student works through a chosen path, the agent provides targeted support---obstacle diagnosis when stuck, validation of intermediate steps, and gentle redirection when the approach reaches a dead end---without prematurely revealing the final answer.
    \item \textbf{Comparative Synthesis}: After solutions are reached, the agent facilitates structured comparison across methods: which is more elegant, which generalizes better, which is more efficient for exam settings. This develops mathematical maturity and strategic flexibility.
    \item \textbf{Variation and Extension}: The profile's output format includes a mandatory ``Variation Extension'' stage where the agent generates related problems by modifying conditions (e.g., changing parameters, relaxing constraints, reversing the problem), building a connected problem network that deepens understanding.
\end{enumerate}

This multi-path approach is particularly effective in mathematics education, where profiles for topics such as analytic geometry and function analysis explicitly require agents to present algebraic, geometric, and calculus-based perspectives for the same problem. The skill modules attached to these profiles provide domain-specific solution templates that the agent can draw upon.

\subsubsection{Assessment Alignment and Standards Integration}

Profiles maintain rigorous alignment with formal assessment standards through the \textit{Standards} section, ensuring that agent interactions prepare students for real-world evaluations. This alignment operates at three levels:

\begin{itemize}
    \item \textbf{National Curriculum Standards}: Each profile references the specific curriculum standards relevant to its subject and grade level (e.g., China's Mathematics Curriculum Standards for Senior High School). The standards section maps profile dimensions to curriculum objectives, ensuring comprehensive coverage and preventing pedagogical drift.
    \item \textbf{Examination Format Awareness}: Profiles for exam-oriented subjects encode knowledge of formal assessment formats---question types, scoring rubrics, time allocation strategies, and common examination pitfalls. For instance, a high school physics profile includes guidance on structured answer formatting that matches the national college entrance examination (Gaokao) requirements.
    \item \textbf{Process-Oriented Evaluation}: Beyond outcome correctness, profiles instruct agents to evaluate and provide feedback on the student's reasoning process. This includes assessing logical coherence, notation correctness, and argument completeness---skills that are increasingly weighted in modern assessment frameworks. The agent's feedback follows a structured rubric encoded in the profile: problem understanding, strategy selection, execution accuracy, and reflection quality.
    \item \textbf{Formative Assessment Integration}: Profiles encode formative assessment checkpoints within the interaction flow. At key junctures, the agent poses diagnostic questions to gauge understanding before proceeding, implementing the ``assess-then-teach'' cycle recommended by~\citep{vanlehn2011relative} as a hallmark of effective tutoring systems.
\end{itemize}

\subsubsection{Cognitive Load Management}

Profiles address cognitive load through structured information presentation strategies:

\begin{itemize}
    \item \textbf{Chunked Presentation}: Complex topics are decomposed into manageable segments. Profiles specify maximum information density per response turn and require explicit comprehension checks between segments.
    \item \textbf{Multi-Modal Representation}: Profiles encourage agents to present information through multiple representations---verbal explanations, symbolic expressions, and references to visual diagrams---reducing the cognitive burden on any single processing channel. The \texttt{tools} field enables agents to invoke diagram generators and symbolic computation tools to support this.
    \item \textbf{Worked Example Fading}: For procedural knowledge, profiles implement a graduated transition from complete worked examples to partially completed examples to independent practice, systematically shifting cognitive effort from studying to doing.
\end{itemize}

\subsubsection{Affective and Motivational Design}

Recognizing that learning is not purely cognitive, profiles encode affective support strategies:

\begin{itemize}
    \item \textbf{Growth Mindset Framing}: Profiles instruct agents to praise effort and strategy rather than innate ability, and to frame challenges as opportunities for growth rather than indicators of inadequacy.
    \item \textbf{Productive Struggle Calibration}: Rather than intervening at the first sign of difficulty, profiles specify wait-time thresholds and escalation conditions, allowing students to experience productive struggle before receiving assistance.
    \item \textbf{Interest Cultivation}: Subject-specific profiles include instructions for connecting abstract concepts to real-world applications and student interests. For example, a physics profile might relate projectile motion to sports scenarios, while a mathematics profile connects probability to game design.
\end{itemize}


\section{Discussion}\label{sec:discussion}

\subsection{Tool Scaling Law (Future Work)}

We hypothesize a complementary \textbf{Tool Scaling Law}: as the repository of callable tools available to an educational agent grows, the agent's actionable capability increases in a predictable manner. Tools in this context include equation solvers, diagram generators, code executors, assessment rubric evaluators, and curriculum databases.

The Tool Scaling Law would formalize the relationship:
\begin{equation}\label{eq:tool-scaling}
    C_{\text{tool}} \propto g\!\left(n_{\text{tools}},\; q_{\text{tools}},\; d_{\text{integration}}\right)
\end{equation}

\noindent where $n_{\text{tools}}$ is the number of available tools, $q_{\text{tools}}$ captures tool quality and reliability, and $d_{\text{integration}}$ measures the depth of integration between tools and the agent's reasoning process.

Planned experiments include: (1) systematically varying the tool set available to agents while holding profiles constant, (2) measuring task completion rates across tool configurations, and (3) identifying critical tool thresholds for different educational domains.

\subsection{Skill Scaling Law (Future Work)}

We further hypothesize a \textbf{Skill Scaling Law}: as domain-specific skill modules deepen and specialize, educational agent expertise scales in a structured manner:

\begin{equation}\label{eq:skill-scaling}
    C_{\text{skill}} \propto h\!\left(n_{\text{skills}},\; d_{\text{depth}},\; c_{\text{composition}}\right)
\end{equation}

\noindent where $n_{\text{skills}}$ is the skill count, $d_{\text{depth}}$ measures the pedagogical depth of individual skills, and $c_{\text{composition}}$ captures emergent capabilities from skill combinations.

The skill repository (1,100+ modules) provides a foundation for investigating this law. Key questions include: How does skill count affect domain coverage? Is there a minimum skill depth threshold for effective tutoring? Do certain skill combinations produce super-additive effects?

\subsection{Interaction Effects}

The three scaling axes---Agent, Tool, and Skill---are not independent. We anticipate interaction effects:

\begin{itemize}
    \item \textbf{Agent $\times$ Tool}: A richer agent profile may enable more effective tool utilization, as the agent has a clearer understanding of when and how to invoke tools.
    \item \textbf{Agent $\times$ Skill}: Profile structure determines how effectively skills are composed and applied; a well-structured profile extracts more value from the same skill set.
    \item \textbf{Tool $\times$ Skill}: Some skills may require specific tools to be effective (e.g., a geometry skill benefits from a diagram tool), creating multiplicative scaling opportunities.
    \item \textbf{Three-Way Interaction}: The full educational agent capability may exhibit emergent properties when all three axes are scaled simultaneously, analogous to how model, data, and compute interact in classical scaling laws.
\end{itemize}

Understanding these interactions will be critical for developing optimal scaling strategies for educational AI systems.

\subsection{Cross-Agent Collaboration for Complex Educational Tasks (Future Work)}

Current educational agents in EduClaw operate primarily as independent specialists, each handling a specific subject or pedagogical function. However, real-world educational tasks are often inherently cross-disciplinary and multi-faceted, requiring expertise that no single agent can provide alone. We envision a future in which heterogeneous educational agents \textit{collaborate} to jointly accomplish complex tasks that exceed any individual agent's capability.

Consider a project-based learning scenario where a student designs a sustainable city. This task simultaneously involves mathematical modeling (optimizing resource allocation), physics reasoning (energy system design), geography knowledge (terrain and climate analysis), and language skills (writing a proposal report). Rather than relying on one general-purpose agent, the platform could orchestrate a \textit{team} of specialized agents---each contributing its domain expertise while coordinating through a shared task context. The AgentProfile specification already supports this vision through the \texttt{subagents} field, which enables hierarchical agent composition. Extending this to peer-level collaboration requires additional mechanisms:

\begin{itemize}
    \item \textbf{Shared Task Context}: A common representation of the learning task, student progress, and intermediate results that all participating agents can read and update, ensuring coherent and non-redundant interactions.
    \item \textbf{Role Negotiation}: Protocols for agents to dynamically determine which agent should lead at each stage of a multi-step task, based on the domain expertise required by the current sub-problem.
    \item \textbf{Pedagogical Consistency}: Mechanisms to ensure that collaborating agents maintain a unified pedagogical stance---consistent scaffolding levels, shared awareness of the student's knowledge state, and aligned assessment criteria---even when they specialize in different subjects.
    \item \textbf{Conflict Resolution}: Strategies for handling situations where agents offer contradictory guidance (e.g., a physics agent and a mathematics agent suggesting different modeling approaches), turning such disagreements into productive learning opportunities for the student.
\end{itemize}

This multi-agent collaboration paradigm would unlock a new dimension of educational capability: just as a team of human teachers with complementary expertise can guide students through ambitious interdisciplinary projects, a team of well-profiled educational agents could provide holistic, coordinated support that scales with the number and diversity of agents in the ecosystem. We believe this represents a natural and promising extension of the Agent Scaling Law, where capability scales not only with the richness of individual profiles but also with the breadth and quality of inter-agent collaboration.

\subsection{Limitations}

Several limitations of the current work should be acknowledged:

\begin{enumerate}
    \item \textbf{Quantitative Evaluation}: Our empirical observations are qualitative. A rigorous validation of the Agent Scaling Law requires controlled experiments with quantitative metrics of educational effectiveness (e.g., learning gains, engagement, pedagogical appropriateness scores).
    \item \textbf{Longitudinal Tracking}: The current platform lacks mechanisms for long-term student progress monitoring, which would be necessary to measure the sustained impact of profile richness on learning outcomes.
    \item \textbf{Multimodal Interaction}: The system is primarily text-based, limiting its applicability to subjects requiring visual or interactive modalities (geometry diagrams, chemistry simulations, physical education demonstrations).
    \item \textbf{Confounding Factors}: Disentangling the effects of profile richness from prompt engineering quality, base model capability, and skill module quality remains challenging.
    \item \textbf{Generalization}: While we demonstrate coverage across K-12 subjects, the scaling law's applicability to other educational contexts (higher education, professional training, informal learning) requires further investigation.
\end{enumerate}


\section{Conclusion}\label{sec:conclusion}

This paper proposed the \textbf{Agent Scaling Law} for educational AI: the principle that educational agent capability scales with the structural richness of its specification---including role definition clarity, pedagogical dimension depth, skill composition, and multi-agent orchestration---not merely with the size of the underlying language model. We operationalized this scaling law through the \textbf{AgentProfile} specification, a structured JSON-based format that enables systematic, declarative definition of educational agents.

We presented \textbf{EduClaw}, a profile-driven multi-agent platform that implements the Agent Scaling Law through an automated construction pipeline, process lifecycle management, and a comprehensive skill repository. The deployment of 330+ agent profiles with 1,100+ skill modules across K-12 subjects provides empirical evidence that structured profile richness correlates with agent capability, supporting the proposed scaling law.

We further identified two complementary scaling axes---the \textbf{Tool Scaling Law} and the \textbf{Skill Scaling Law}---as future research directions, and discussed the potential interaction effects among all three axes. Together, these scaling laws suggest that the path to more capable educational AI lies in building \textit{stronger structured capability systems}, not solely in training larger models.

Future work will focus on: (1) rigorous quantitative validation of the Agent Scaling Law through controlled experiments, (2) empirical investigation of Tool and Skill Scaling Laws, (3) extension to multimodal and collaborative learning scenarios, (4) development of optimization strategies that balance scaling across all three axes for maximum educational impact, and (5) enabling cross-agent collaboration, where heterogeneous educational agents with complementary expertise cooperate to jointly accomplish complex, interdisciplinary educational tasks that exceed the capability of any single agent.

\section*{Acknowledgments}

We gratefully acknowledge the authoritative education experts---including experienced K-12 teachers, curriculum designers, and educational researchers---who provided essential pedagogical guidance, curriculum alignment validation, and quality assurance throughout the construction of the educational skill library and agent profiles. Their domain expertise was instrumental in ensuring that the platform's educational capabilities meet professional standards. We also thank all contributors to the skill repository for their sustained efforts in building and refining the pedagogical knowledge base.

\bibliographystyle{plainnat}
\bibliography{ref}

\end{CJK}
\end{document}